# Human-level Performance On Automatic Head Biometrics In Fetal Ultrasound Using Fully Convolutional Neural Networks

Matthew Sinclair, Christian F. Baumgartner, Jacqueline Matthew, Wenjia Bai, Juan Cerrolaza Martinez, Yuanwei Li, Sandra Smith, Caroline L. Knight, Bernhard Kainz, *Senior Member, IEEE,* Jo Hajnal, Andrew P. King, Daniel Rueckert, *Fellow, IEEE*

*Abstract*— Measurement of head biometrics from fetal ultrasonography images is of key importance in monitoring the healthy development of fetuses. However, the accurate measurement of relevant anatomical structures is subject to large inter-observer variability in the clinic. To address this issue, an automated method utilizing Fully Convolutional Networks (FCN) is proposed to determine measurements of fetal head circumference (HC) and biparietal diameter (BPD). An FCN was trained on approximately 2000 2D ultrasound images of the head with annotations provided by 45 different sonographers during routine screening examinations to perform semantic segmentation of the head. An ellipse is fitted to the resulting segmentation contours to mimic the annotation typically produced by a sonographer. The model's performance was compared with inter-observer variability, where two experts manually annotated 100 test images. Mean absolute model-expert error was slightly better than inter-observer error for HC (1.99mm vs 2.16mm), and comparable for BPD (0.61mm vs 0.59mm), as well as Dice coefficient (0.980 vs 0.980). Our results demonstrate that the model performs at a level similar to a human expert, and learns to produce accurate predictions from a large dataset annotated by many sonographers. Additionally, measurements are generated in near real-time at 15*fps* on a GPU, which could speed up clinical workflow for both skilled and trainee sonographers.

## I. Introduction

Accurate assessment of fetal development is of paramount importance to ensure the continued wellbeing of mothers and newborns both during and after pregnancy. A mid-trimester ultrasonography (US) scan, typically carried out between 18-22 weeks gestation, is used in most countries as part of standard prenatal care. Standard plane views are acquired during the scan, in which distinct anatomical features can be identified [1]. Measurements taken of the head, abdomen and femur are commonly used to estimate both fetal age and weight. Additionally, when measured at different points in time, such biometrics can be used to assess fetal growth trajectory and ensure normal fetal development. However, early detection rates of fetal abnormalities are low and vary greatly depending on geographical location. This is largely due to the high level of skill required for sonographers to navigate to the predefined standard image planes and to subsequently measure standard biometrics [2].

Additionally, ultrasound images suffer from a range of pitfalls including acoustic shadow, motion blurring and low signal-to-noise ratio, making the identification of standard planes a challenging task for sonographers. Furthermore, once the standard planes have been identified, there is considerable inter-observer variability in the measurement of different anatomical structures depending on level of sonographer expertise and directed attention fatigue. Caliper placement variation between sonographers is the largest source of error in fetal biometric measurements accounting for up to 80%, more than the error attributed to differences in fetal orientation or patient anatomy [2].

### A. Related Work

Recent work has demonstrated the ability of convolutional neural networks (CNNs) to be robust to the varied imaging conditions of US for standard plane classification [3]. For biometric estimation of head circumference (HC) and biparietal diameter (BPD) from US, several (non-deep learning) methods have been proposed [4, 5, 6], none of which however have demonstrated human-level performance and run in real-time. Additionally, more recent work has proposed the use of a fully convolutional network (FCN) for the automatic segmentation of the fetal head, demonstrating improved results using a cascaded network architecture [7]. In that study however no comparison was made to inter-expert variability, nor were biometrics estimated.

### B. Contributions

In this work, we propose an approach to measure HC and BPD from the standard trans-ventricular (TV) brain view plane. A FCN is trained using almost 2000 clinically annotated images, more than double any previous work using a deep network for fetal head segmentation [7]. Subsequently an ellipse is fitted to the predicted segmentation contours to mimic the measurement procedure used by sonographers. We assess the performance of our method by comparing to intra- and inter-expert errors with a 100 patient subset of the test data, demonstrating human-level performance of the proposed method. We achieve lower error and variance, and much faster (near real-time) inference compared to other automated approaches for biometric estimation [4, 5, 6], which also saves the 20*s* per scan typically required for manual annotation.

M. Sinclair, W. Bai, J. Cerrolaza Martinez, Y. Li, B. Kainz and D. Rueckert were with the Biomedical Image Analysis Group, Department of Computing, Imperial College London, London SW7 2AZ, U.K. (e-mail: m.sinclair@imperial.ac.uk).

C. Baumgartner was with the Computer Vision Lab, ETH Zurich.

J. Matthew, C. Knight, S. Smith, J. Hajnal and A. P. King were with the School of Biomedical Engineering and Imaging Sciences, King's College London, London SE1 7EH, U.K.,

J. Matthew and C. Knight were also with the Biomedical Research Centre, Guy's and St Thomas' NHS Foundation, London SE1 9RT, U.K.

## II. MATERIALS

The study population consists of 2,724 2D ultrasound examinations from volunteers at 18-22 weeks gestation, which were acquired and labeled during routine screening by a team of 45 expert sonographers according to UK FASP guidelines [1]. All volunteers gave written informed consent in accordance with ethical committee approval. Eight different ultrasound systems, all of the same make and model (GE Voluson E8), were used to perform the examinations. Each volunteer was examined by a single sonographer, who made patient-specific adjustments to probe settings (e.g. dynamic range, power, thermal index, etc.) to identify all standard scan planes including the TV plane. From the TV plane, the sonographer created an annotation of the fetal head diameter during the examination using an ellipse tool available on the ultrasound system. Freeze-frame DICOM images of the annotated head in the TV planes were saved for all subjects. Screen-capture videos were also saved for the ultrasound examinations, which on average consisted of 13 minutes of footage per subject at a frame rate of 30*fps*.

While each examination is performed according to the FASP guidelines, large variability is still introduced by the variations in acoustic shadow artifacts, selected probe settings, zoom level, fetal anomalies and the position of the fetus, which determines the image content around the head. No special selection was made to remove cases with particular artifacts or anatomical anomalies from the dataset. Thus, our dataset is a good representation of the variety of TV view images one might expect to see in a clinical setting.

## III. METHODS

### A. Data Preprocessing

The image preprocessing steps are listed in Table 1. First, ellipse annotations in the freeze-frame DICOM images were extracted to provide the ground-truth head region. A dashed line indicates the location of the expert-defined ellipse, and was identified by the unique RGB value relative to the underlying US image. An ellipse was fitted to the dashed lines to produce a mask of the head region, serving as the ground truth segmentation. Pixels inside the ellipse were given the value 1, and those outside the ellipse 0. A comparison of the fitted ellipse diameter to that recorded on the US system indicated a match with <0.2% error.

To obtain annotation-free images, the videos were parsed for the matching unannotated frame corresponding to each DICOM image. This was achieved by (1) performing OCR to identify the time-stamp of the frame (which becomes frozen once the sonographer has identified the standard plane for annotation); (2) identifying the frame immediately before any annotation is created in the image, but after any adjustments to the image have been made (e.g. changes in zoom and orientation by the user). A visual check was performed to ensure all recovered video frames were unannotated and matched the corresponding annotated DICOM image for all subjects. Images were cropped to remove on-screen text and scaled by 0.5 to a size of 320x384 pixels. The mean original pixel size was 0.13x0.13mm, so down-sampling introduced an error of up to about 0.26mm in BPD and 0.82mm in HC measurements, or approximately 0.5% of the mean values.

TABLE I. DATA PREPROCESSING STEPS

| Step | Data Source |
|---|---|
| 1) Generate ellipse mask image | DICOM |
| 2) Retrieve unannotated frame | Video feed |
| 3) Crop and scale images | DICOM and video feed |

### B. Convolutional Neural Networks

Deep learning has produced state-of-the-art results in many computer vision and medical image analysis problems, including semantic segmentation [8]. Briefly, CNNs designed for semantic segmentation such as the FCN [8] architecture learn a set of image filters at multiple spatial scales, producing hierarchical feature maps of increasing coarseness. Further filters then learn to up-sample the coarse feature maps to produce a pixel-wise label prediction at the resolution of the input image. A FCN with 16 convolutional layers is trained to segment the head region in the TV plane images. Fig. 1 illustrates the full network architecture.

Formally, let $x$ be an unannotated image of the head, and $y$ the corresponding ground-truth pixel-wise label map from the extracted ellipse mask. A training set $S = \{x_i | i = 1, 2, ..., N; y_i | i = 1, 2, ..., N\}$ consists of pairs of images and label maps. Supervised learning is performed to estimate the network parameters, $\Theta$, to predict label map $y_i$ of image $x_i$ in the training set, by optimising the cross-entropy loss,

$$\min_\Theta L(\Theta) = -\sum_i \sum_j \log P(y_{i,j} | x_i, \Theta), \quad (1)$$

where $j$ denotes the pixel index and $P(y_{i,j} | x_i, \Theta)$ denotes the softmax probability produced at pixel $j$ for image (and label map) $i$. An argmax operation on the softmax probability map is used to produce a binary segmentation, which is then used to compute HC and BPD (see *Section C*).

Figure 1. The VGG-16 FCN architecture. Convolution filters are denoted with [filter size]/[stride] (e.g. 3x3/1), and up-conv filters up-sample feature maps with the necessary scale factor to produce same *x* and *y* dimensions as the input image. Number of filters indicated above feature maps.

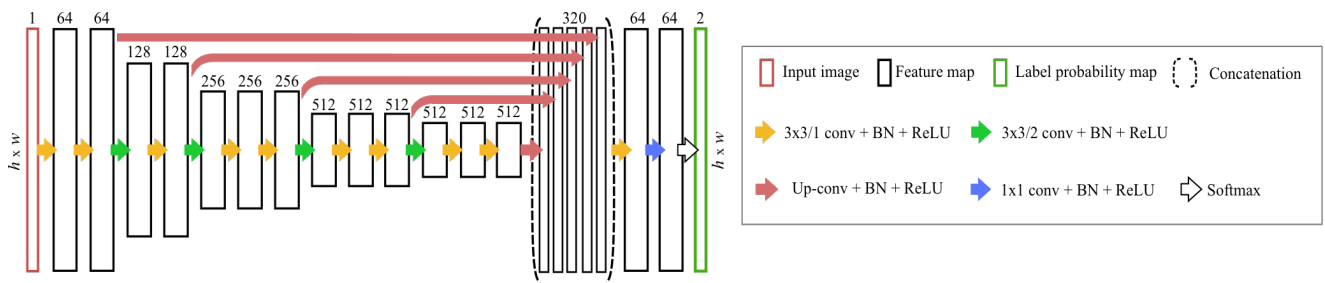

## C. Biometric Estimation from Segmentation

To estimate HC and BPD, an ellipse is fitted [9] to the contours of the FCN segmentation. Ellipse parameters include centroid coordinates, $c_x$ and $c_y$, major and minor axis radii, $a$ and $b$, all in pixels, and angle of rotation, $\alpha$. The biometrics BPD and HC are estimated as follows:

$$\text{BPD} = b\, s_{xy}, \quad (2)$$

where $s_{xy}$ is the isotropic pixel size in millimeters of the input image. HC is estimated using the Ramanujan approximation II [10],

$$HC \approx \pi\,(a+b)\left(1 + \frac{3h}{10+\sqrt{4-3h}}\right) s_{xy}, \quad (3)$$

$$h = \frac{(a-b)^2}{(a+b)^2}.$$

The error of this approximation is $O(h^{10})$, which for more circular ellipses like those for the fetal head is negligible [10].

## IV. EXPERIMENTS AND RESULTS

### A. Experiments

**Network Training:** The image data was randomly separated into train/test splits of 80%/20%, and then the train split was further separated into a train/validation split of 90%/10%, resulting in train, test and validation sets of 1948, 539 and 216 images, respectively. The FCN, initialized with weights pre-trained for ImageNet classification, was trained for up to 20 epochs or until there was no improvement in validation Dice for 3 epochs. The Adam optimizer with a constant learning rate of 1e-5 was used. A mini-batch size of 5 images was used, with random left-right flipping for data augmentation. After training, inference was performed on the test set and biometrics were estimated with Eq.'s (2) and (3).

**Intra- and Inter-observer study:** In addition to assessing the model on the 539-image test set annotated by 45 expert sonographers, the model was also compared to intra- and inter-observer variability on 100 randomly selected test images. Two experts (*expert 1:* an engineer with substantial ultrasound experience, and *expert 2:* a trained sonographer) used the ellipse tool in MITK[1] to generate two annotations for each image. The images were randomly sampled from the 100-image set until each image was seen twice by both experts. Biometrics were computed from the ellipse annotations using Eq.'s (2) and (3). Mean absolute error (MAE) and mean error (ME) were used to compare intra-expert, inter-expert, and model-expert performance for estimating HC and BPD, and Dice Coefficient was used to assess ellipse overlap.

Inter-expert error was computed for each image from the differences between each of *expert 1*'s measurements and each of *expert 2*'s measurements (i.e. 4 differences per image). Model-expert error was computed from the differences between the model-derived measurement and each expert's measurements (i.e. also 4 differences per image). Mean and standard deviation (SD) of all metrics were computed across the 100-sample set annotated by the two experts, as well as for the original annotations on the whole 539-image test dataset.

[1] The Medical Imaging Toolkit (MITK), website: mitk.org

TABLE II. RESULTS ON 100 TEST IMAGES AND ON ENTIRE 539-IMAGE TEST SET (RIGHT-MOST COLUMN). VALUES SHOWN ARE MEAN (SD).

| Metric | Intra-expert 1 | Intra-expert 2 | Mean Inter-expert | Mean Model-expert | All test data |
|---|---|---|---|---|---|
| **HC MAE (mm)** | 1.55 (1.30) | 1.55 (1.14) | 2.16 (1.16) | 1.99 (0.87) | 1.80 (1.49) |
| **HC ME (mm)** | 0.18 (2.01) | -0.09 (1.92) | 1.56 (1.70) | 1.01 (1.62) | 0.54 (2.28) |
| **BPD MAE (mm)** | 0.40 (0.32) | 0.60 (0.45) | 0.59 (0.34) | 0.61 (0.33) | 0.68 (0.62) |
| **BPD ME (mm)** | -0.05 (0.51) | -0.06 (0.75) | 0.01 (0.60) | 0.29 (0.55) | 0.13 (0.91) |
| **Dice Coefficient** | 0.983 (0.006) | 0.984 (0.005) | 0.980 (0.005) | 0.980 (0.005) | 0.981 (0.007) |

### B. Results

Table II shows the results of the tests comparing expert performance and that of the proposed model, as well as the performance on the entire test set. Naturally the intra-expert MAEs are generally lowest. Inter-expert Dice and BPD errors are comparable to model-expert Dice and BPD errors. However, the inter-expert MAE and ME are slightly higher than the model-expert error for HC. This suggests that the model is producing ellipse annotations that lie very close to, or even in between, the two experts' annotations. This could be attributed to the fact that the network is trained on a big dataset annotated by a large number of sonographers, which may reduce the bias of its prediction relative to any individual expert.

The errors on the entire test set (annotated by 45 sonographers) are also comparable to the model-expert errors for the 100-image set. Fig. 2 illustrates results on the whole test set, including the worst cases out of the 539 images in terms of Dice and absolute error for HC and BPD. HC and BPD ME±SD on all test data are 0.54±2.28mm and 0.13±0.91mm, respectively, outperforming reported results of the winning entries of the ISBI 2012 fetal segmentation challenge [4] (HC: -2.01±3.29mm, BPD: 0.58±1.24mm), and a more recent study using Random Forests and fast ellipse fitting [5] (HC: 1.70±5.29mm).

Figure 2. Ellipses generated from the model (red) versus ground truth annotations (yellow) from test set images. The test images with the largest errors for HC, Dice and BPD are shown in the right-most column.

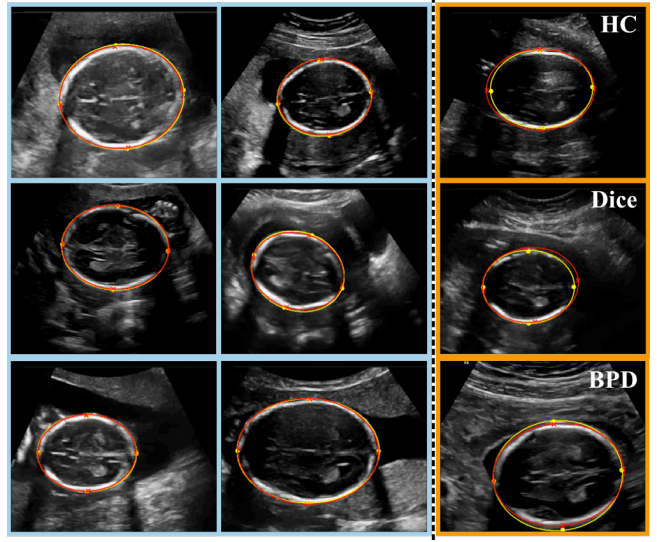

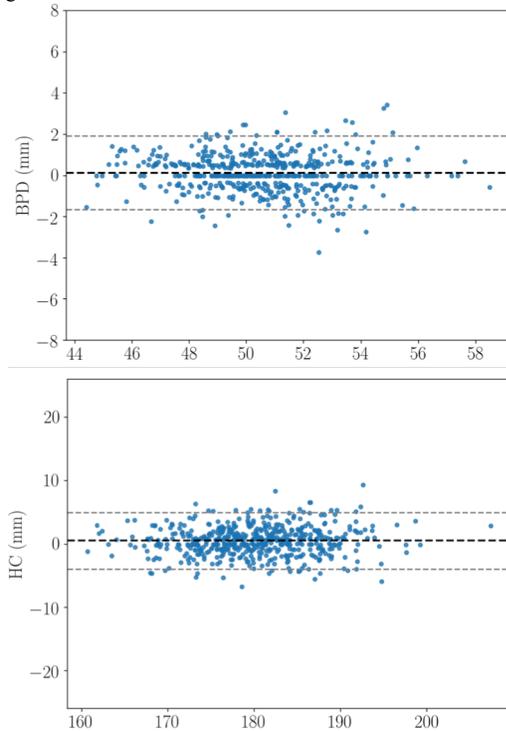

Figure 3. Bland-Altman Plots for HC and BPD on all test data.

Importantly, our method does not produce any major outlier predictions even in the varied 539 images of the test set as shown in the Bland-Altman plots in Fig. 3 and also demonstrated in Fig. 2. This indicates the robustness of our method to the wide range of image conditions seen in clinical fetal ultrasound. Additionally, inference with the proposed method runs at 15*fps* on a workstation with a Nvidia Titan Xp GPU, allowing for effective real-time assessment of fetal biometrics during an examination. This saves not only the time required for a sonographer to create an annotation which typically takes about 20*s*, but also allows for a continuous examination, saving time required to reorient the probe after pausing to create an annotation.

Compared to a recent deep-learning approach for fetal head segmentation [7], the Dice Coefficient of our model on the test set of 0.980 is better than the Dice achieved with a similar architecture in [7] of 0.966, possibly due to the considerably larger dataset used for the training of our model (1948 *vs* 900 images). In [7] Dice was increased to 0.984 with the use of a cascaded network, although the annotations were all created by a single expert and were not ellipses but rather segmentations with boundaries that closely follow the outer skull. This means that the networks in [7] were more likely to learn the particular bias of the single expert unlike in this study. Furthermore, the inter-expert Dice of ellipse annotations is 0.980 in this study, which is slightly higher than that reported in [4] (0.978).

## V. Conclusion

We have presented a deep learning approach to estimate fetal head biometrics from a standard TV plane, which is routinely acquired at 20 weeks gestation to assess fetal health and development. To the best of our knowledge, this is the first method proposed for this purpose that both produces human-level performance and operates in near real-time. The proposed method is trained on a larger training set than any we have found in the literature and produces low-bias ellipses, leveraging training annotations from 45 experts. Our method has applications not only for clinical screening, but can be especially helpful for training sonographers and use in settings lacking clinical experts.

Of course, accurate estimation of fetal biometrics first requires the identification of standard planes, which recent work has demonstrated can be greatly assisted by CNNs [3]. In future work, CNNs can be extended to the quantification of biometrics from other standard views, including the abdominal and femur views which are also important for assessment of fetal growth. Additionally, the network can be adapted to work with images from other ultrasound systems via methods such as fine-tuning or domain adaptation.

## Acknowledgments

This work was supported by the Wellcome Trust IEH Award [102431, iFind]. The authors acknowledge financial support from the Department of Health via the National Institute for Health Research (NIHR) comprehensive Biomedical Research Centre award to Guy's & St Thomas' NHS Foundation Trust in partnership with King's College London and King's College Hospital NHS Foundation Trust. The authors also thank Nvidia Corporation for the donation of a Titan Xp GPU used in this study.